\documentclass[letterpaper]{article} 
\usepackage{multirow}
\usepackage{booktabs}
\usepackage{siunitx, makecell, graphicx}    
\usepackage{aaai2026} 
\usepackage{times}  
\usepackage{helvet}  
\usepackage{courier}  
\usepackage[hyphens]{url}  
\usepackage{graphicx} 
\usepackage{natbib}  
\usepackage{caption} 
\usepackage{algorithm}
\usepackage{algorithmic}
\usepackage{amsmath}
\usepackage{newfloat}
\usepackage{listings}
\DeclareCaptionStyle{ruled}{labelfont=normalfont,labelsep=colon,strut=off} 
\floatstyle{ruled}
\newfloat{listing}{tb}{lst}{}
\floatname{listing}{Listing}
\title{Kunlun Anomaly Troubleshooter: Enabling Kernel-Level Anomaly Detection and Causal Reasoning for Large Model Distributed Inference}
\author{
    Yuyang Liu\textsuperscript{\rm 1}\equalcontrib\thanks{This work was supported by Alibaba Group through Alibaba Innovative Research Program}
    Jingjing Cai\textsuperscript{\rm 1}\equalcontrib,
    Jiayi Ren\textsuperscript{\rm 2}\equalcontrib,
    Peng Zhou\textsuperscript{\rm 2},
    Danyang Zhang\textsuperscript{\rm 2},
    Yin Du\textsuperscript{\rm 2},
    Shijian Li\textsuperscript{\rm 1}
}
\affiliations{
    \textsuperscript{\rm 1}  Zhejiang University, Hangzhou, China\\
    \textsuperscript{\rm 2} Alibaba Cloud, Hangzhou, China\\
    *\{yuyangliu, caijj\}@zju.edu.cn, renjiayi@alibaba-inc.com

}

\begin{document}

\maketitle

\begin{abstract}
Anomaly troubleshooting for large model distributed inference (LMDI) remains a critical challenge. Resolving anomalies such as inference performance degradation or latency jitter in distributed system demands significant manual efforts from domain experts, resulting in extremely time-consuming diagnosis processes with relatively low accuracy. In this paper, we introduce Kunlun Anomaly Troubleshooter (KAT), the first anomaly troubleshooting framework tailored for LMDI. KAT addresses this problem through two core innovations. First, KAT exploits the synchronicity and consistency of GPU workers, innovatively leverages function trace data to precisely detect kernel-level anomalies and associated hardware components at nanosecond resolution. Second, KAT integrates these detection results into a domain-adapted LLM, delivering systematic causal reasoning and natural language interpretation of complex anomaly symptoms. Evaluations conducted in Alibaba Cloud Service production environment indicate that KAT achieves over 0.884 precision and 0.936 recall in anomaly detection, providing detail anomaly insights that significantly narrow down the diagnostic scope and improve both the efficiency and success rate of troubleshooting.
\end{abstract}

\section{1 Introduction}

\begin{figure*}[h]
\centering
\includegraphics[width=0.8\textwidth]{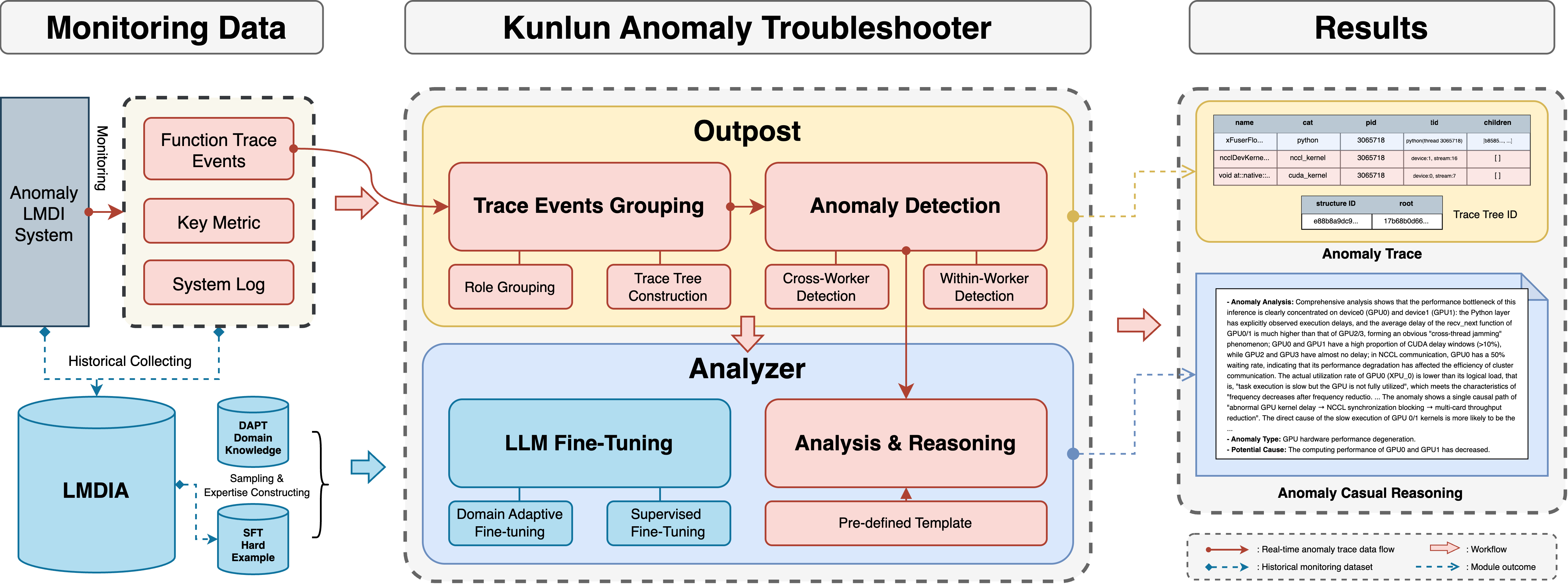}
\caption{Overview of Kunlun Anomaly Troubleshooter}
\label{fig：kat_overview}
\end{figure*}

Distributed computing systems have advanced significantly, enabling the implementation of increasingly large models which now reach billions of parameters and demand tremendous computation power~\cite{brown2020languagegpt3, dubey2024llama}. To meet these intensive computational requirements, large model distributed inference (LMDI) systems usually adopt parallel computing techniques~\cite{mpi, dagum1998openmp, chowdhery2023palm}, resulting in complex architectures integrating numerous interconnected software and hardware components. However, this increasing exposes systems to a higher likelihood of anomalies, such as task failures, latency jitter, and performance degradation. Such anomalies significantly diminish user experience and pose critical operational challenges for cloud service providers~\cite{hagemann2020adforcloud}.

\textbf{Troubleshooting of anomalies in LMDI systems} presents unique challenges that distinct from traditional anomaly detection scenarios. Due to the large number of software and hardware components and their intricate interdependencies within LMDI systems, anomalies can originate from diverse sources and propagate across multiple system stack~\cite{deng2025minder, liu2021microhecl, wu2020microrca}. It usually fails to accurately locate root causes merely identifying anomalous machines or independently analyzing isolated metrics, as they cannot reveal the underlying causal relationships among anomaly symptoms. Furthermore, current troubleshooting practices for distributed system rely heavily on manual inspection and correlation across extensive software and hardware stacks. Given the sheer volume and variety of anomaly symptoms, such manual analyses require significant domain expertise and usually overlook critical anomaly information, resulting in misjudgment of the root cause with unbearable time-consuming. Therefore, an effective LMDI troubleshooting approach must satisfy two essential capabilities: (1) \textbf{precise identification of workload functions and hardware components impacted by anomalies}; and (2) \textbf{explicit reasoning of the causal chains underpinning these anomalies.}

To provide these capabilities, we first construct \textbf{LMDIA}, a full-stack multimodal dataset comprising 42 large-model inference tasks from 11 challenging LMDI anomaly scenarios in production. The dataset consists of 3 modalities including function trace events, providing a full-stack system coverage from bottom hardware component up to high-level model service and profound details about the system behaviors at kernel function granularity with nanosecond resolution. Based on LMDIA, we present Kunlun Anomaly Troubleshooter (\textbf{KAT}), the first anomaly troubleshooting framework tailored for LMDI which provides accurate anomaly detection on kernel function trace events, and delivers anomaly causal reasoning leveraging the rich kernel-level insights. Unlike traditional supervised or unsupervised approach which employs a baseline or normal pattern comparison method, KAT exploits the synchronicity and consistency of parallel computing GPU workers to perform anomaly detection on the kernel function trace events. Furthermore, KAT integrates a domain-specific LLM finetuned to systematically correlate anomalies across software behaviors and hardware states, delivering interpretable anomaly reasoning in nature language. The reasoning results specifically reveal the causal chain of the anomaly symptoms and help engineer to locate the root cause. KAT implements these capabilities through two complementary modules: \textbf{Outpost} for precise anomaly detection, and \textbf{Analyzer} for interpretable causal reasoning.

In summary, we make the following contributions:
\begin{itemize}
    \item We construct LMDIA, a full-stack multimodal datasets based on the most challenging LMDI anomaly cases.
    \item We propose the KAT, the first LMDI anomaly troubleshooting methodology which performs anomaly detection at nanosecond resolution and kernel-level granularity, and delivers interpretable root cause analysis at the same time.
\end{itemize}

\section{2 Background and Challenge}
\subsection{2.1 Relative Works and Limitations}

Conventional Time-Series Anomaly Detection (TSAD)~\cite{zamanzadeh2024dltsad} methods typically rely on supervised or unsupervised approaches, utilizing models such as GNN~\cite{ding2023mstgat}, LSTM, AE~\cite{audibert2020usad, zhang2024tracecontrast}, or Attention modules~\cite{tranad} to learn normal behavioral patterns from historical data, and detect anomalies via forecasting or reconstruction-based loss analysis. Particularly in cloud microservice systems~\cite{soldani2022adrcaformicro, soldani2022cloudtsadsurvey1}, these approaches leverage telemetry data composed of API traces~\cite{xie2023gtrace}, hardware metrics~\cite{wang2023coral}, and system logs~\cite{meng2019loganomaly}. However, such approaches are unsuitable in the context of LMDI anomaly troubleshooting. First, the fundamental assumption of stable and repeatable operational patterns underlying traditional TSAD methods is invalid in highly variable LMDI environments, where workloads and configurations can vary dramatically. Additionally, telemetry data sampled at second-level granularity inherently lacks sufficient temporal resolution, resulting in the loss of critical anomaly details in inference tasks typically lasting tens of seconds.

Recent studies have started addressing anomaly detection within distributed systems. Minder \cite{deng2025minder} utilizes machine-level similarity to detect anomalies on hardware metrics but is limited to the detection on fault machines and LM training scenario. C4 \cite{dong2024boostingc4} primarily targets predictable GPU communication latency issues prevalent in parallel training tasks. Moreover, these methods overly emphasize hardware conditions, neglecting the critical role of software-level insights in anomaly localization and troubleshooting. Several works have introduced time-series modalities into LLM for anomaly detection and analysis~\cite{wang2025chattime, timer}, but they are typically constrained by the limited detection accuracy~\cite{tan2024arellmactually, merrill2024llmstruggletotsad}. To the best of our knowledge, we are the first work that uses kernel-level function trace data for anomaly detection and causal reasoning analysis in distributed systems.

\subsection{2.2 Function Trace Data}
The trace data we adopt in this paper is the detailed function execution records down to kernel-level, which significantly differ from API call records commonly collected in microservice works~\cite{guo2020gmtamicrotrace, zhang2022tracecrl}. 

\begin{figure}[h]
\centering
\includegraphics[width=1\columnwidth]{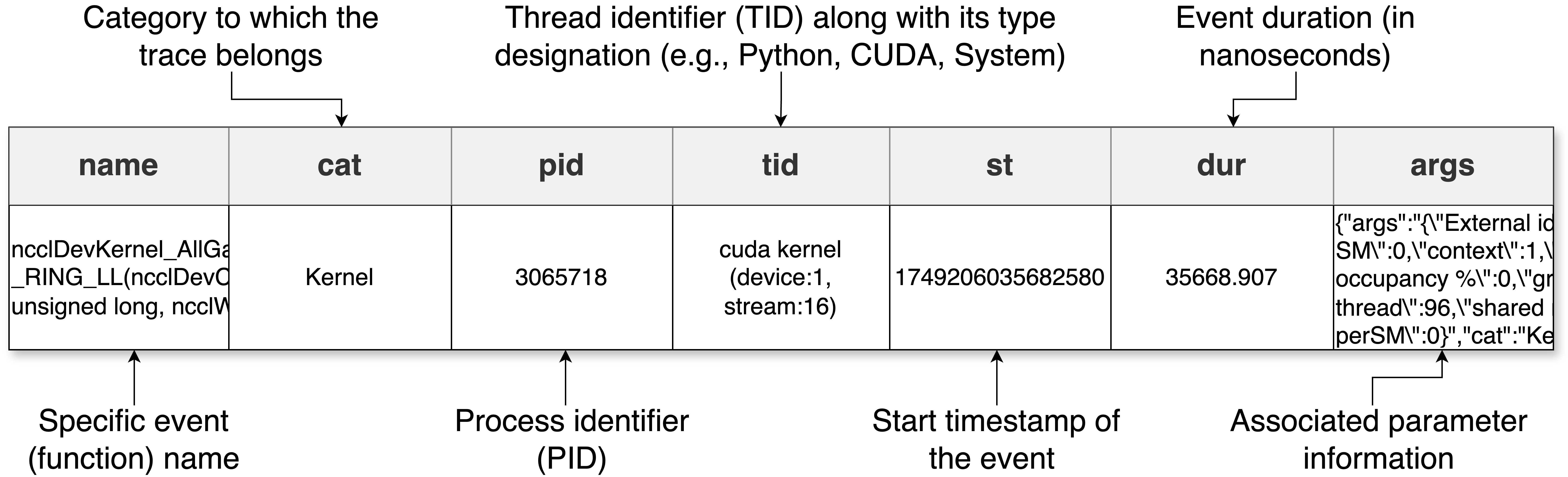}
\caption{Example of Trace Event}
\label{fig：trace_example}
\end{figure}

As shown in figure \ref{fig：trace_example}, a function trace typically contains several attributes such as function name, PID, TID, execution duration, and various contextual arguments. Trace events provide detailed insights at the function call level, precisely capturing the system state and timing dependencies among software and hardware components. They enable deep analysis of spatiotemporal characteristics and propagation patterns of anomalies at nanosecond resolution. Therefore, trace event data not only supports the precise detection of anomalous kernel functions and impacted hardware components but also elucidates the underlying mechanism of anomaly propagation, establishing it as a crucial basis for automated anomaly detection and root cause analysis.

\subsection{2.4 Challenges}
The primary challenges we need to address are summarized as follows:

\textbf{C1: Data Availability and Granularity Constraints.}
One major barrier to troubleshoot anomaly in LMDI is the lack of robust \textit{data monitoring framework} and open-source \textit{fine-grained datasets}. Existing monitoring data typically consist of hardware metrics sampled per second and high-level system logs, lacking both the detailed software behaviors (e.g., kernel executions, function call dependencies) and the high temporal resolution which are essential for precise anomaly detection and root cause analysis.

\textbf{C2: Anomaly Detection on Trace Event.}
Extracting feature from trace events is quite challenging due to their discrete nature, high information density and large volume. Additionally, LMDI is inherently \textit{task-driven}, exhibiting variations in data distribution patterns. Any change of inference configuration like inference frameworks or optimization parameters would result in a new data distribution, making it impractical to learn a “normal” pattern from historical data for every LMDI anomaly case.

\textbf{C3: Complex Anomaly Symptoms for Causal Reasoning.}
Performance anomalies originating from a single component in distributed system usually manifest across multiple interdependent monitoring data channels, exhibiting a propagation patterns over time dimension. Locating the root cause typically requires experts to synthesize anomaly symptoms from diverse data channels to make a comprehensive analysis. However, the numerous hardware and software components involved in LMDI tasks, along with their complex inference workflows, significantly increase diagnostic complexity. 

\section{3 Methodology}

\subsection{3.1 Opportunities}
Through extensive experimentation and detailed analyses of anomaly cases observed in production environments, we find two key insights that form the basis of KAT :

\textbf{Synchronicity and consistency of parallel worker.} The wide adoption of distributed inference frameworks such as DeepSpeed and vLLM~\cite{kwon2023vllm} have standardized parallel computing paradigms that allocate computational and communication tasks across processes and devices in a coordinated manner. This coordination results in workers executing identical tasks exhibit notable synchronicity and consistency in their trace events. Furthermore, the frequent reuse of fundamental network architectures and operators in LLMs leads to a repetition of kernel execution within single worker. Consequently, kernel execution trace events exhibit both cross-worker consistency from parallel scheduling and strong temporal regularity within an individual inference session.

\begin{figure}[h]
\centering
\includegraphics[width=1.0\columnwidth]{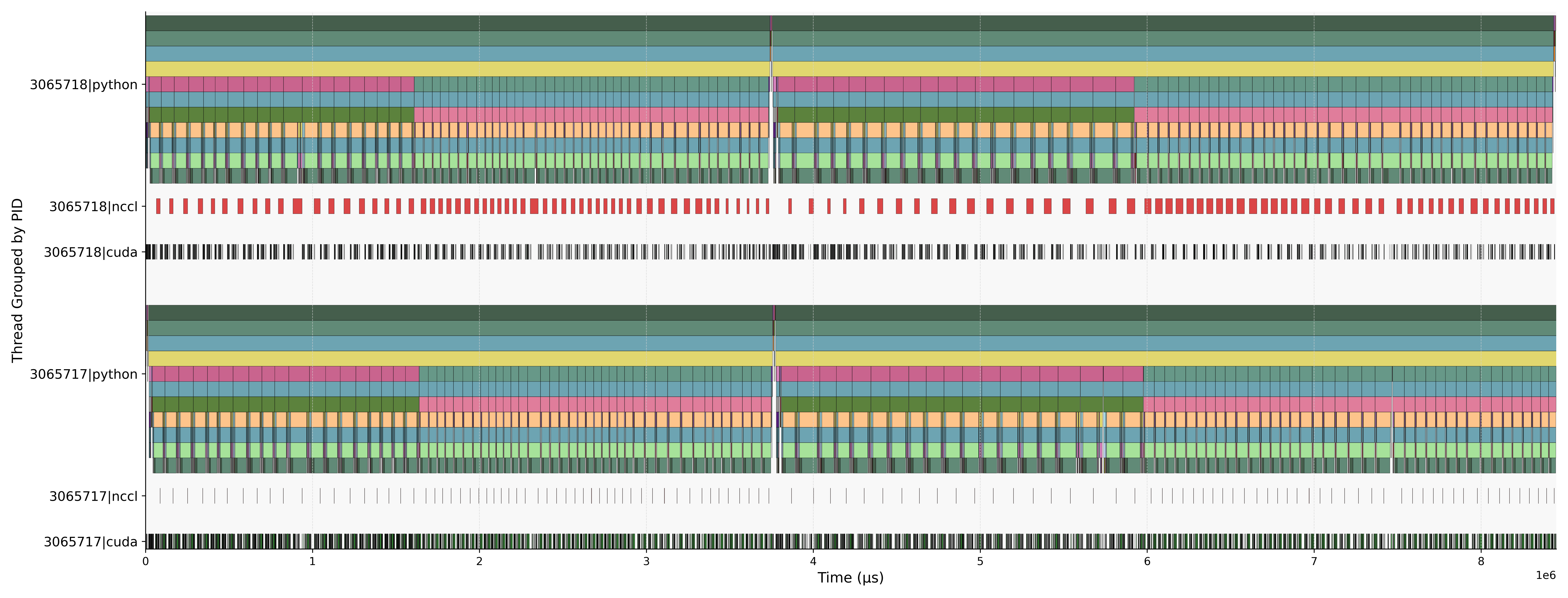}
\caption{Trace Timelines of a Two Parallel GPUs Inference}
\label{fig：c&s_of_trace}
\end{figure}

\textbf{LLM reasoning capability.} There has been a growing number of works looking into employing LLMs instead of traditional deep learning models in the areas of TSAD and root cause localization (RCL) for microservice and cloud systems~\cite{chen2024rcacopilot, guan2025dabl, roy2024exploringagentforrca}. These approaches embed monitoring data into the semantic space of LLM to provide anomaly detection and cause analysis expressed in natural language. Leveraging LLM strong powerful semantic understanding and context awareness abilities, we can reveal the cause chain of anomalies, helping engineering efficiently locate the root cause. 

\begin{figure*}[htb]
\centering
\includegraphics[width=0.9\textwidth]{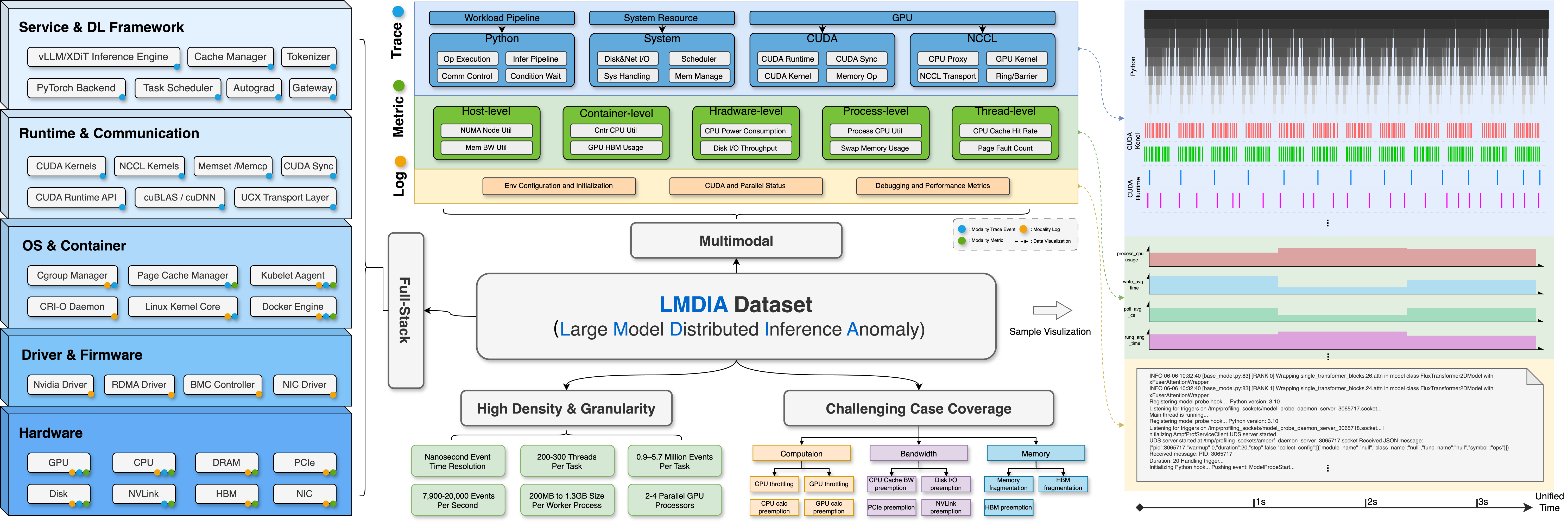}
\caption{LMDIA Data Monitoring Structure}
\label{fig:dataset_strucutre}
\end{figure*}

\subsection{3.2 Design Overview}
We first present a fine-grained, full-stack dataset containing kernel-level function trace events. Building upon this dataset, we propose KAT, a system that decouples anomaly detection from causal reasoning. This two-module design allows the detection algorithm to be specifically tailored to the characteristics of LMDI scenarios, boosting its accuracy. Concurrently, the reasoning module focuses exclusively on critical anomalies to provide precise and in-depth root cause analyses. As illustrated in Figure \ref{fig：kat_overview}, we have made efforts in three aspects:
\begin{itemize}
    \item \textbf{Full-stack Multimodal Dataset} (section 3.3): We construct LMDIA, a monitoring dataset covering full-stack of LMDI system and 3 data modalities. The dataset is collected on challenging LMDI anomaly cases and uniquely introduces kernel function trace event modality. 
    \item \textbf{Trace Event-based Anomaly Detection} (section 3.4): We design Outpost, an anomaly detection module employs the synchronicity and consistency of parallel GPU workers, providing anomaly detection results on trace events at kernel-level granularity and nanosecond resolution.
    \item \textbf{LLM-based Root Cause Analysis} (section 3.5): We further develop Analyzer, a domain-specialized large language model (LLM) trained via domain-adaptive pre-training (DAPT)~\cite{gururangan2020dapt} and supervised fine-tuning (SFT)~\cite{ouyang2022sft} on hard examples experts craft~\cite{ahmed2023domainftllm0, prabhakar2025omniscience}. Analyzer integrates anomaly events detected by Outpost with supplementary monitoring data into a pre-defined instruction template, generating interpretable anomaly causal analyses.
    
\end{itemize}

\subsection{3.3 Full-stack Dataset Construction}
We introduce LMDIA, a comprehensive full-stack multimodal dataset constructed from 42 independent LM inference tasks spanning 3 anomaly types including 11 representative anomaly scenarios in production environments. 
As illustrated in Figure~\ref{fig:dataset_strucutre}, the dataset stands out prominently in two distinctive characteristics: (1) \textbf{Full-stack System Coverage}. LMDIA captures system behaviors spanning low-level hardware operations, CUDA runtime events~\cite{nickolls2008cuda}, NCCL collective communication~\cite{li2019nvlink}, and high-level Python model inference logic. And the monitoring data encompasses three complementary modalities: trace events, performance metrics, and system logs. This comprehensive view enables holistic understanding of complex interactions in distributed inference workflows. (2) \textbf{High Information Density and Granularity}. Each monitored LMDI anomaly case involves 200-300 threads in total, capturing approximately 0.9–5.7 million events per task at rates ranging from 7,900 to 20,000 events per second. Individual trace file sizes range from 200MB to 1.3GB per worker process, providing nanosecond-level visibility into detailed software-hardware interactions.

To provide broad applicability, LMDIA categorizes anomalies into 3 primary types: computational degradation, bandwidth congestion, and storage and memory resource constraints. These are further subdivided into 11 specific subtypes, such as GPU frequency throttling and system resource contention. The diversity of the dataset is reinforced by including representative LMDI tasks from diverse domains, including NLP and text-to-image generation~\cite{devlin2019bert, rombach2022stablediffusion}. A detailed summary of key dataset characteristics is available in the Appendix.

\subsection{3.4 Anomaly Detection on Trace Events}
Anomalies in distributed inference mainly manifest as performance degradation or system failures. Performance degradation, typically due to misconfiguration or hardware preemption, is difficult to diagnose precisely, as it slows inference without producing explicit errors or logs. Based on the insights into trace events discuss in Section 2.2, we develop Outpost to detect anomalies from function execution perspective. Unlike conventional historical baseline method, Outpost employs the synchronicity and consistency of parallel GPU workers and adopt a training-free statistical comparison approach. 

\begin{figure*}[htb]
\centering
\includegraphics[width=0.9\textwidth]{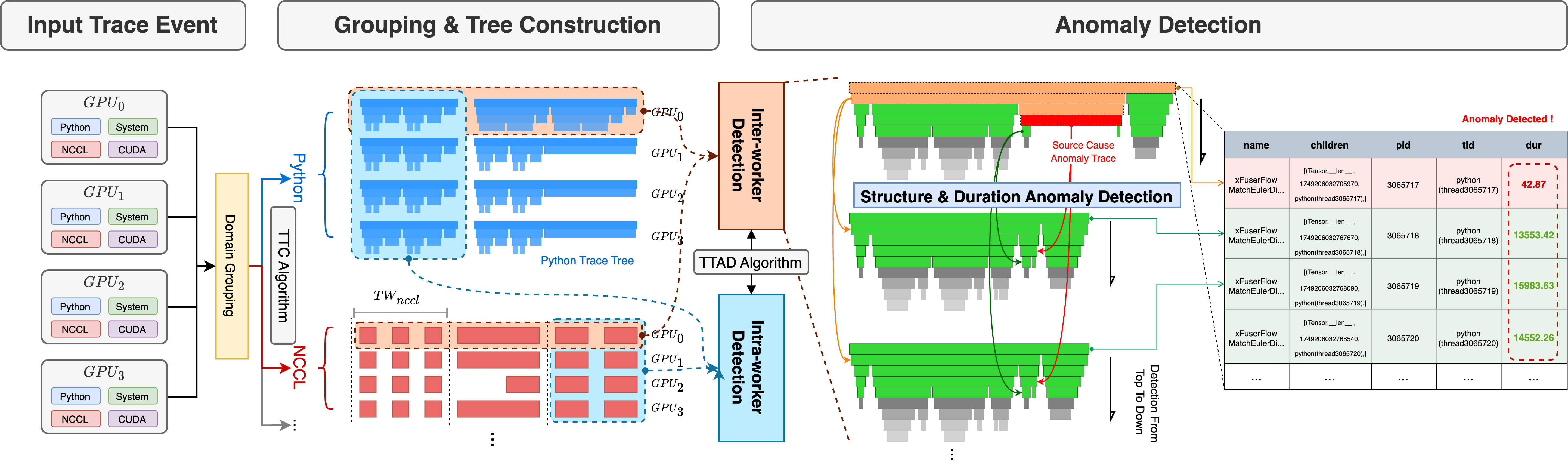}
\caption{Pipeline of KAT Outpost Anomaly Detection on Trace Data}
\label{fig：ad_pipeline}
\end{figure*}

\subsubsection{3.4.1 Trace Grouping and Trace Tree Construction.} 
\textbf{Role-based Grouping.} The objective of trace grouping is to clearly delineate distinct execution contexts across parallel processes and threads, thereby ensuring precise anomaly detection by comparing behaviors of trace events corresponding to the same parallel task. In typical distributed inference frameworks, concurrent tasks are assigned for multiple GPU workers, each corresponding to a unique process ID (PID). Each worker further launches various threads (identified by TID), which can be categorized by their domain roles, such as Python scheduling threads, CUDA kernel execution threads, and NCCL communication threads. Based on this observation, Outpost first group trace events by their PIDs, and subsequently subdivide them by TIDs. Finally, these events are categorized into different thread roles domain. This hierarchical grouping strategy efficiently organizes trace data according to their execution contexts, providing distinct and comparable objects for subsequent anomaly detection.

\begin{algorithm}[htb]
\small
\caption{Trace Tree Construction (TTC)}
\label{alg:ttc}
\textbf{Input}: $T^d = \{t^d_0, t^d_1, ... ,t^d_n\}$: trace events within a thread $d$.\\
\textbf{Output}: $(Root, Cildren)$: all root nodes and children nodes of each trace event.
\begin{algorithmic}[1] 
\STATE $\tilde{T}^d \gets BioTimeSort(\;T^{d}\;)$ by $(st\,\uparrow,\;et\,\downarrow)$
\STATE $Root \gets \{\ \}$ root node dict
\STATE \(Children \gets\) \(\{\,t.id : [\ ] \mid t \in \tilde{T}^d\,\}\) children node dict
\STATE $Stack \gets [\ ]$ trace events stack
\FOR{$t^d_i$ in $\tilde{T}^d$}
    \WHILE {$\neg \; Stack.empty\; \land \;Stack.top.et \; \le \; t^d_i.st$\;}
        \STATE $Stack.\mathrm{pop}()$
    \ENDWHILE
    \IF {$\neg\; Stack.empty$}
        \STATE $Root[\;t^d_i.id\;] \gets Stack.top.id$
        \STATE $Cildren[\;Stack.top.id\;].\mathrm{append}(\;t^d_i.id, \;t^d_i.st \;)$
    \ELSE
        \STATE $Root[\;t^d_i.id\;] \gets null$
    \ENDIF
    \STATE $Stack.\mathrm{push}(\;t^d_i\;)$
\ENDFOR
\STATE \textbf{return} $(Root, \;Cildren)$
\end{algorithmic}
\end{algorithm}

\textbf{Hierarchical Trace Trees.} Trace events of python threads typically exhibit nested function invocation structures, which indicates essential dependencies among function executions. To leverage this structural information, Outpost build hierarchical structures called trace trees to precisely extract the intrinsic parent-child relationships among events~\cite{xie2023gtrace, zhang2022deeptralog}. Within each Python thread, trace events naturally exhibit nested intervals defined by their start timestamp ($st$) and duration ($dur$). Formally, given two trace events $t_i$ and $t_j$, event $t_j$ is a child (direct or indirect) of $t_i$ if its interval lies entirely within that of $t_i$. By iteratively applying this relationship, Outpost constructs an accurate trace tree for each Python thread group. To ensure structural consistency across trees, each constructed trace tree is normalized and encoded using SimHash. Structurally similar trace trees, determined by pairwise Hamming distances below a threshold $\sigma$, will be assigned a common structural identifier ($str\_id$). In contrast, events from CUDA, NCCL and system threads are typically independent kernel events without nested structures. To maintain consistency in data structure for subsequent detection, these single-layer kernel executions are treated as leaf nodes or root nodes without child nodes, each forming an independent flat trace event structure.

\subsubsection{3.4.2 Anomaly Detection Methodology}
Most performance issues in LMDI originate from minority of components and persist consistently throughout an entire inference task, and these anomalies would not exhibit variation over time. Besides, variability of workloads and inference configurations introduces complexity in data distribution as discussed in Challenge 2. These two distinctive characteristics undermine traditional anomaly detection methods because such approaches typically rely on predefined stable baselines derived from historical data, which fail when data distributions dynamically shift or when anomalies remain stable. Meanwhile, anomalies may still concurrently arise across parallel workers during inference. For instance, when a data center triggers power-capping, causing all GPUs to simultaneously drop to a lower P-State, the execution duration of kernels within the same inference stage consistently doubles across all parallel workers. In such cases, comparisons across workers become ineffective.

Motivated by these unique characteristics and challenges, we propose the Trace Tree Anomaly Detection (TTAD) shown in Algorithm \ref{alg:ttad}. TTAD leverages consistency and synchronicity across parallel GPU workers to dynamically compare runtime behaviors captured in kernel-level function trace events. Specifically, the anomaly detection is conducted from two complementary dimensions:

\begin{algorithm}[ht]
\small
\caption{Trace Tree Anomaly Detection (TTAD)}
\label{alg:ttad}
\textbf{Input}: $\mathcal{TT}_r=\{TT_0,\dots,TT_n\}$: trace data to detect\\
\textbf{Parameter}: $\lambda > 0$: significance coefficient\\
\textbf{Output}: $\mathcal{A} = \{a_0, a_1, \dots, a_k\}$
\begin{algorithmic}[1]
  \STATE $\mathcal{A} \gets [\;]$,  
         $evt \gets [\; TT_0.\mathrm{root},\; \dots,\; TT_n.\mathrm{root}\; ]$
  \WHILE{$evt \neq [\;]$}
    \STATE $H \gets [\;]$
    \FORALL{$e \in evt$}
      \IF{$e.\mathrm{name}\notin H$}
        \STATE $H.\mathrm{append}(e.\mathrm{name})$
        \STATE $G\gets\{\;w\in\ evt \mid w.\mathrm{name}=e.\mathrm{name}\;\}$
        \STATE $(\mu_g,\sigma_g)\gets\mathrm{MeanAndStd}(\{w.\mathrm{dur}\mid w\in G\})$
        \FORALL{$w\in G$}
          \IF{$|w.\mathrm{dur} - \mu_g|  \ge \lambda\sigma_g$}
            \STATE $\mathcal{A}.\mathrm{append}(w)$
          \ENDIF
        \ENDFOR
      \ENDIF
    \ENDFOR
    \STATE $evt \gets [\;e_n \mid e_n\in e.\mathrm{children},\; \forall e \in evt\;]$
  \ENDWHILE
  \RETURN $\mathcal{A}$
\end{algorithmic}
\end{algorithm}

\textbf{Inter-worker detection} identifies anomalies by analyzing structural and temporal consistency across parallel GPU workers. These anomalies typically manifest as structural discrepancies or significant variations in execution durations across parallel workers performing identical computational tasks. TTAD traverses the trace trees layer-by-layer, grouping trace events by their structural identifier ($str_id$) and function name. Within each group, we apply statistical analysis based on event durations. Formally, for events at layer $l$ belonging to function $f$, we calculate the sample mean $\mu_{l,f}$ and standard deviation $\sigma_{l,f}$ of their durations. A trace event $w$ is flagged as anomalous if its duration deviates from the mean by more than $\lambda$ standard deviations: $|\mathrm{dur}(w) - \mu_{l,f}| \geq \lambda \sigma_{l,f}$. Because of the dependencies among function calls, anomalies detected at upper layers often propagate downward. TTAD thus iteratively identifies anomalies from top to bottom layers, isolating the deepest anomalous event as the initial trigger or root cause.

\textbf{Intra-worker detection} focuses on identifying anomalies by examining repetitive execution patterns and temporal deviations within individual worker processes. Unlike inter-worker detection, which compares behaviors across parallel workers, intra-worker detection leverages repetitive execution patterns within the same worker. Specifically, TTAD identifies structurally identical trace trees generated repeatedly within each worker thread and applies the same statistical detection method introduced in inter-worker detection to analyze event durations. By performing such intra-worker analyses, TTAD effectively detects anomalies caused by internal or global resource constraints, which inter-worker comparisons alone may fail to capture.

\subsection{3.5 Anomaly Causal Reasoning and LLM Training}
We further develop the domain-specific LLM Analyzer to provide comprehensive causal reasoning for anomalies after obtaining all anomaly trace events detected by Outpost. The analysis consists anomaly symptoms summarization and a critical anomaly causal chain~\cite{wei2022cotreasoning} which enable engineers to accurately pinpoint the root cause, ultimately facilitating effective troubleshooting of LMDI anomalies. Analyzer’s reasoning ability primarily derived from two key components: (1) a well-defined instruction template that structurally encapsulates critical anomaly context and information, and (2) domain-adaptive pre-training that equips the model with essential domain-specific knowledge, further enhanced by supervised fine-tuning on expert-crafted hard examples to strengthen its causal reasoning capability.

\subsubsection{3.5.1 Instruction Template}
\begin{figure}[htbp]
\centering
\includegraphics[width=1\columnwidth]{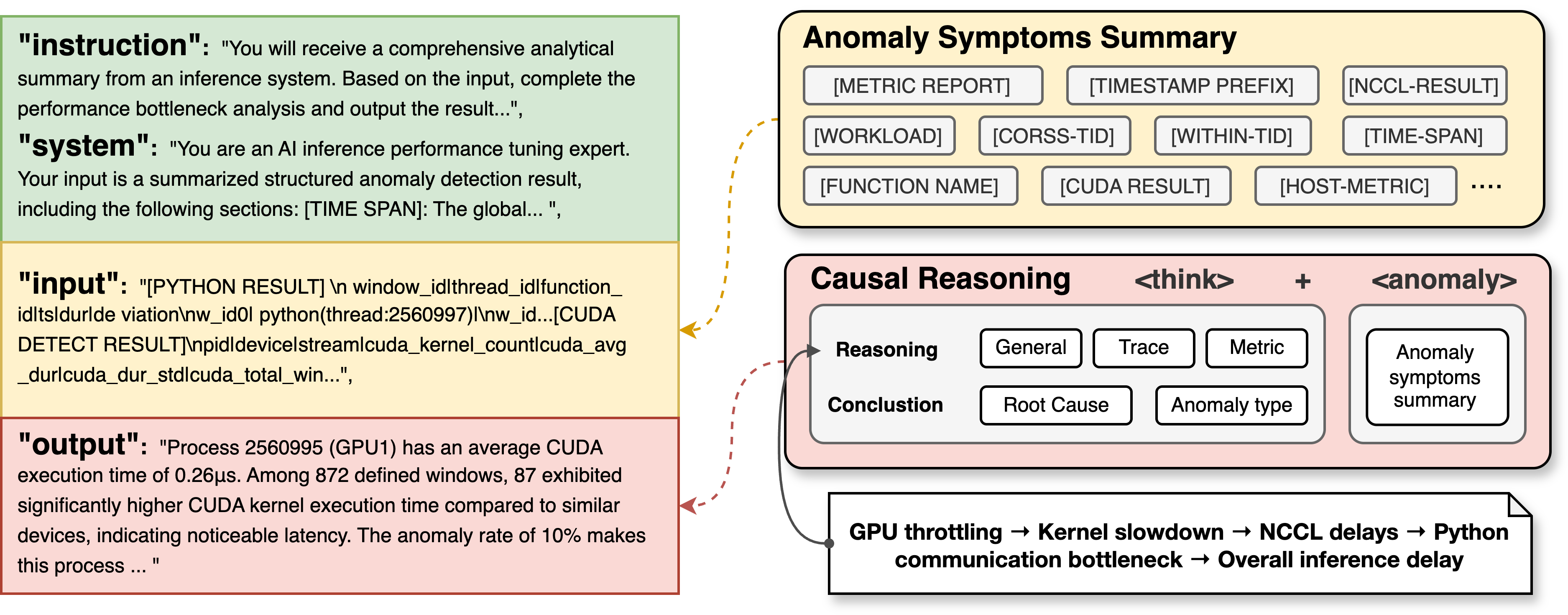}
\caption{Instruction Template for SFT Hard Example}
\label{fig:prompt_template}
\end{figure}

To enable the LLM to fully leverage anomaly events detected by Outpost, we construct an instruction template that structurally organize all anomaly symptoms and other supporting data into clear semantic segments. As illustrated in Figure \ref{fig:prompt_template}, the ``input" section comprises three parts: inference task configurations, all anomaly trace events and other metric data summary. Specifically, we list anomalous events according to thread domains, explicitly retaining the original function names. These names inherently provide rich contextual information, including the type of computational task performed, the inference execution stage involved, and the hardware resources or software libraries utilized during their execution. Besides, although the metric data sampled per second may be relatively sparse, it can still serve as supplementary information to aid anomaly analysis.

\subsubsection{3.5.2 Model Fine-tuning}
Analyzer is fine-tuned based on Qwen-14B-Instruct. First, we perform DAPT to integrate essential LMDI domain-specific knowledge into the base model and get Analyzer-Base, which serves as a robust semantic foundation. The data for pretraining contains 12 relevant technical domains such as CUDA libraries and container technology, and subsequently cleaned through Data-Juicer~\cite{chen2024datajuicer}. 

Guided by the principle of prioritizing data quality over quantity, we construct 26 high-quality manually annotated \textbf{hard examples} to further perform SFT on Analyzer. These carefully curated examples are collected from representative LMDI anomaly cases, each annotated by multiple senior domain experts. Each training sample comprises a summary of anomaly symptoms, a structured causal analysis, and a summarized conclusion. Notably, experts carefully annotate a detailed and professional \textbf{anomaly causal chains} in analysis that effectively bridges observed anomalies and domain knowledge. For instance, if one GPU triggers frequency throttling due to hardware issues, this slowdown cascades through the multi-GPU inference process, causing kernel execution delays, prolonged NCCL wait times, and ultimately increased latency in critical Python-level communication functions. These high-quality expert insights and reasoning steps are integrated into the “output” section using clearly defined tags (e.g., \textless think\textgreater). This structured approach addresses the complexity introduced by anomaly propagation patterns in LMDI, significantly improving the effectiveness of engineers in locating root causes.

\begin{table*}[t]
\centering
\small            
\caption{Performance comparison on causal reasoning task. Metrics include textual quality (ROUGE-L, Valid Format Rate), causal reasoning accuracy and quality ($\Delta$RCA Precision, Recall, F1-score, Entailment Score), result consistency (BERTScore, EmbeddingSim, Jaccard), and subjective evaluation by LLM-Judge (GPT-4o calibrated Faithfulness, Answer Relevancy).}
\label{tab:llm_perf}

\begin{tabular}{@{}
    c                 
    l                 
    | S[table-format=1.3]
      S[table-format=1.3]   
    | S[table-format=1.3]
      S[table-format=1.3]
      S[table-format=1.3]
      S[table-format=1.3]   
    | S[table-format=1.3]
      S[table-format=1.3]
      S[table-format=1.3]   
    | S[table-format=1.3]
      S[table-format=1.3]  
  @{}}
\toprule
\multirow{2}{*}{\textbf{Type}}
  & \multirow{2}{*}{\textbf{Model}}
  & \multicolumn{2}{c|}{\textbf{Quality}}
  & \multicolumn{4}{c|}{\textbf{Causal Analysis Quality}}
  & \multicolumn{3}{c|}{\textbf{Self Consistency}}
  & \multicolumn{2}{c}{\textbf{LLM-Judge}} \\
\cmidrule(lr){3-4}\cmidrule(lr){5-8}\cmidrule(l){9-11}\cmidrule(l){12-13}
  & 
  & {R-L} & {VFR}
  & {$\Delta$R-P} & {$\Delta$R-R} & {$\Delta$R-F1} & {EC}
  & {BSc} & {ESim} & {Jac}
  & {Faith} & {AnsRel} \\
\midrule
\multirow{3}{*}{\rotatebox{90}{Flagship}}
  & GPT-4o            &0.235  &0.717 	&0.447 	&0.390 	&0.416 	&0.285 	&0.862 	&0.828 	&0.079 	&0.733 	&0.803 \\
  & Gemini-2.5-Pro    &0.209  &0.495 	&0.359 	&0.233 	&0.282 	&0.268 	&0.860 	&0.644 	&0.047 	&0.600 	&0.770 \\
  & Claude Sonnet 4   &0.235  &0.899 	&\textbf{0.807} &\textbf{0.586} &\textbf{0.674} &\textbf{0.334} &\textbf{0.903}	&0.811 	&0.092 	&0.703 	&0.853  \\
\midrule
\multirow{4}{*}{\rotatebox{90}{Open}}
  & DeepSeek-R1       &0.202  &0.727 	&0.324 	&0.214 	&0.257 	&0.306 	&0.864 	&0.779 	&0.029 	&0.577 	&0.670 \\
  & Kimi-K2           &0.225  &0.737 	&0.239 	&0.173 	&0.200 	&0.225 	&0.867 	&0.812 	&0.083 	&0.661 	&0.844 \\
  & Llama3-70B        &0.221  &0.736  &0.174  &0.069  &0.099  &0.180  &0.861  &0.754  &0.044  &0.373  &0.627 \\
  & Qwen2.5-14B-Ins   &0.243  &0.667 	&0.154 	&0.142 	&0.147 	&0.251 	&0.841 	&0.722 	&0.021 	&0.240 	&0.623 \\
\midrule
\multirow{2}{*}{\rotatebox{90}{KAT}}
  & Analyzer-Base     &0.274  &0.859 	&0.426 	&0.225 	&0.293 	&0.288 	&0.884 	&0.782 	&0.067 	&0.460 	&0.640  \\
  & \textbf{Analyzer} &\textbf{0.290} &\textbf{0.970} &\textbf{0.716} &\textbf{0.360} &\textbf{0.479} &\textbf{0.322} &\textbf{0.894} &\textbf{0.849} &\textbf{0.131} &\textbf{0.763} 	&\textbf{0.903} \\
\bottomrule
\end{tabular}
\end{table*}

\section{4 Experiment}
To comprehensively evaluate the effectiveness of KAT in diagnosing performance anomalies in LMDI, we design two sets of experiments: \textbf{E1:} Performance evaluation of Outpost in detecting anomalous trace events; \textbf{E2:} Effectiveness assessment of Analyzer in root cause analysis.

\subsection{4.1 Dataset and Evaluation Metrics}
\textbf{Dataset.} To evaluate the effectiveness of Outpost in detecting anomalous trace events under challenging scenarios, we use all 42 anomaly samples from the LMDIA dataset, ensuring coverage across diverse and complex anomaly scenarios common in LMDI environments. Due to the data scarcity, we selected three challenging and representative LMDI anomaly samples based on the instruction template for Analyzer evaluation. The “output” section of each evaluation sample is crafted by domain experts, serving as ground truth. These evaluation cases were excluded from the fine-tuning process to ensure an unbiased assessment of the Analyzer anomaly reasoning capability.

\textbf{Metric.} For anomaly trace event detection, we evaluate performance using standard metrics: Precision, Recall, and F1-score. Specifically, correctly identified anomaly events are considered True Positives (TP), missed anomaly events are False Negatives (FN), and normal events incorrectly flagged as anomalies are False Positives (FP). For the root causal reasoning task, we evaluate it from 4 dimensions: output quality, causal reasoning correctness and rationality, result consistency and subjective evaluation. Accordingly, we select metrics including ROUGE-L~\cite{lin2004rougel}, Valid Format Rate~\cite{wang2019vfr}, $\Delta$RCA, Entailment Score~\cite{entailmentscore}, Self Consistency~\cite{wang2022selfconsistency}, and LLM-Judge Score~\cite{zheng2023llmjudging}. Detailed definitions and associated sub-metrics are presented in the evaluation results Table \ref{tab:llm_perf}.

\subsection{4.2 Outpost Performance in Anomaly Detection}
\begin{figure}[htb]
\centering
\includegraphics[width=1.0\columnwidth]{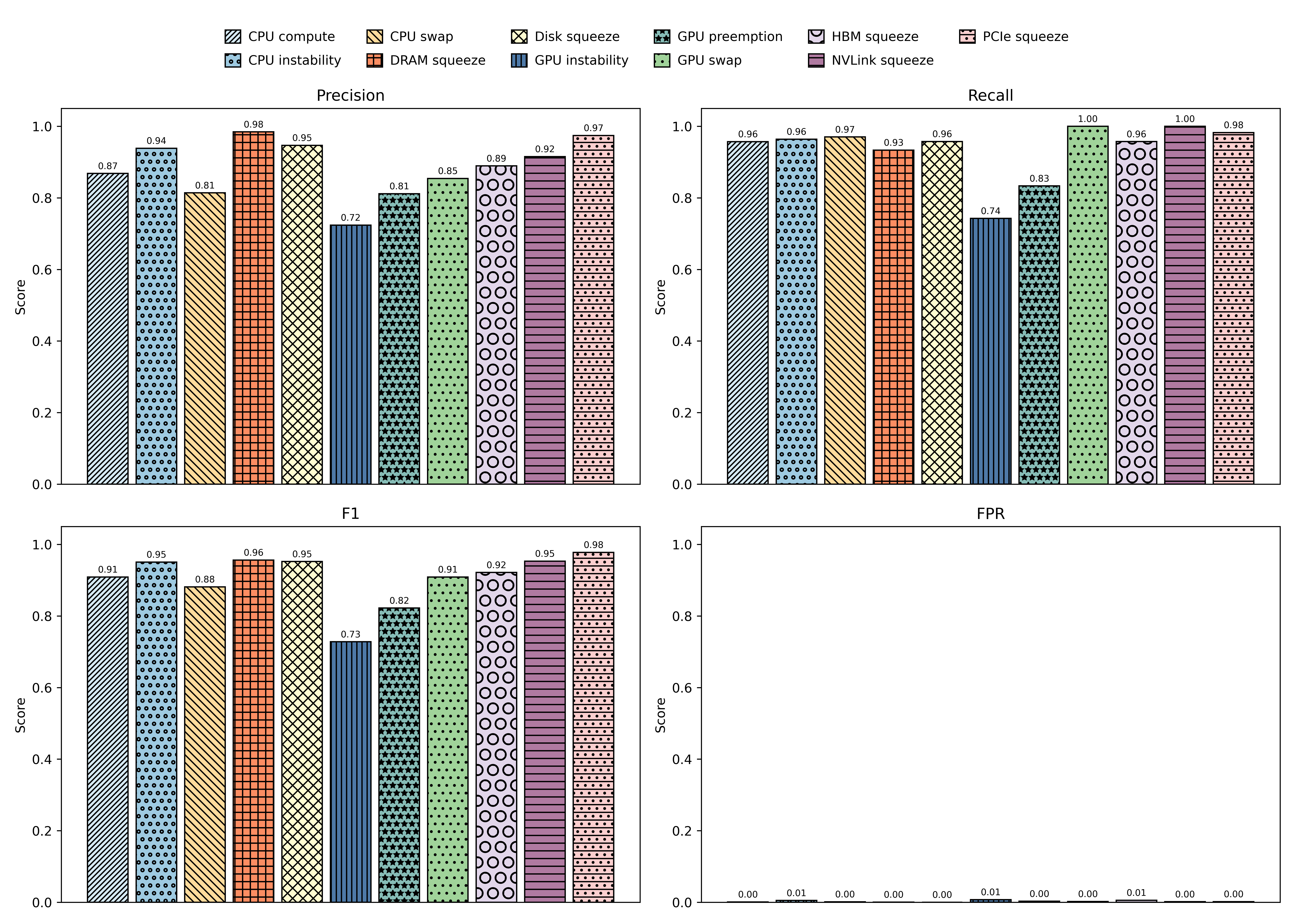}
\caption{Accuracy of Outpost Trace Anomaly Detection}
\label{fig:eval_res_outpost}
\end{figure}

As shown in Figure~\ref{fig:eval_res_outpost}, Outpost achieved 88.4\% precision and 93.6\% recall, resulting in a balanced F1-score of 90.1\% on average. These results demonstrate Outpost’s strong capability in detecting most of anomalous events while minimizing false alarms. Notably, Outpost exhibits a very low false positive rate (FPR) of only 0.27\% which indicates that engineers can confidently rely on detected anomalies as accurate starting points, thus save much troubleshooting time.

\subsection{4.3 Analyzer Effectiveness in Root Cause Analysis}
In this experiment, we systematically evaluate our proposed Analyzer model against state-of-the-art large language models, including leading proprietary models and widely used open-source models. The results in Table~\ref{tab:llm_perf} show that Analyzer exhibits competitive performance on multiple evaluation metrics. It exceeds all evaluated open-source models and notably outperforms several industrial-leading proprietary models including GPT-4o and Gemini-2.5-Pro. For instance, Analyzer achieves significant improvements over open-source baselines, achieving improvements of up to 45.4\% in VFR and up to 365\% in key causal analysis metrics ($\Delta$RCA). Subjective evaluations further validate Analyzer, with improvements of 88.2\% in faithfulness and 28.9\% in answer relevancy of anomaly analysis. These results clearly show effectiveness of Analyzer, indicating that it can accurately locate anomalies and produce anomaly causal reasoning at a level comparable to state-of-the-art proprietary models. 
Nevertheless, we observe minor performance gaps on particular specific causal analysis metrics compared to flagship proprietary models such as Claude 4-sonnet. This suggests inherent model-scale advantages and architectural strengths of proprietary models, as well as potential constraints due to limited domain fine-tuning data for Analyzer. Notably, we observe tremendous performance improvements from Qwen-14B-Instruct to Analyzer-Base, and subsequently from Analyzer-Base to Analyzer. These improvements clearly demonstrates that the specialized domain knowledge from DAPT and the anomaly reasoning abilities derived by expert-curated hard examples through SFT collectively strengthen Analyzer’s domain-specific capabilities.

\section{5 Conclusion}
In this paper, we first develop LMIDA, a comprehensive multimodal dataset comprising representative and challenging anomaly scenarios with full-stack, high-resolution monitoring data. Building on this, we introduce KAT, a novel troubleshooting approach tailored for LMDI anomalies which performs anomaly detection on function trace event and delivering interpretable causal reasoning analyses. 

\bibliography{aaai2026}

@article{mpi,
  title={A high-performance, portable implementation of the MPI message passing interface standard},
  author={Gropp, William and Lusk, Ewing and Doss, Nathan and Skjellum, Anthony},
  journal={Parallel computing},
  volume={22},
  number={6},
  pages={789--828},
  year={1996},
  publisher={Elsevier}
}

@article{dagum1998openmp,
  title={OpenMP: an industry standard API for shared-memory programming},
  author={Dagum, Leonardo and Menon, Ramesh},
  journal={IEEE computational science and engineering},
  volume={5},
  number={1},
  pages={46--55},
  year={1998},
  publisher={IEEE}
}

@article{chowdhery2023palm,
  title={Palm: Scaling language modeling with pathways},
  author={Chowdhery, Aakanksha and Narang, Sharan and Devlin, Jacob and Bosma, Maarten and Mishra, Gaurav and Roberts, Adam and Barham, Paul and Chung, Hyung Won and Sutton, Charles and Gehrmann, Sebastian and others},
  journal={Journal of Machine Learning Research},
  volume={24},
  number={240},
  pages={1--113},
  year={2023}
}

@article{brown2020languagegpt3,
  title={Language models are few-shot learners},
  author={Brown, Tom and Mann, Benjamin and Ryder, Nick and Subbiah, Melanie and Kaplan, Jared D and Dhariwal, Prafulla and Neelakantan, Arvind and Shyam, Pranav and Sastry, Girish and Askell, Amanda and others},
  journal={Advances in neural information processing systems},
  volume={33},
  pages={1877--1901},
  year={2020}
}

@article{dubey2024llama,
  title={The llama 3 herd of models},
  author={Dubey, Abhimanyu and Jauhri, Abhinav and Pandey, Abhinav and Kadian, Abhishek and Al-Dahle, Ahmad and Letman, Aiesha and Mathur, Akhil and Schelten, Alan and Yang, Amy and Fan, Angela and others},
  journal={arXiv e-prints},
  pages={arXiv--2407},
  year={2024}
}

@inproceedings{deng2025minder,
  title={Minder: Faulty machine detection for large-scale distributed model training},
  author={Deng, Yangtao and Shi, Xiang and Jiang, Zhuo and Zhang, Xingjian and Zhang, Lei and Zhang, Zhang and Li, Bo and Song, Zuquan and Zhu, Hang and Liu, Gaohong and others},
  booktitle={22nd USENIX Symposium on Networked Systems Design and Implementation (NSDI 25)},
  pages={505--521},
  year={2025}
}

@inproceedings{liu2021microhecl,
  title={Microhecl: High-efficient root cause localization in large-scale microservice systems},
  author={Liu, Dewei and He, Chuan and Peng, Xin and Lin, Fan and Zhang, Chenxi and Gong, Shengfang and Li, Ziang and Ou, Jiayu and Wu, Zheshun},
  booktitle={2021 IEEE/ACM 43rd International Conference on Software Engineering: Software Engineering in Practice (ICSE-SEIP)},
  pages={338--347},
  year={2021},
  organization={IEEE}
}

@inproceedings{wu2020microrca,
  title={Microrca: Root cause localization of performance issues in microservices},
  author={Wu, Li and Tordsson, Johan and Elmroth, Erik and Kao, Odej},
  booktitle={IEEE/IFIP Network Operations and Management Symposium (NOMS)},
  year={2020}
}

@article{dong2024boostingc4,
  title={Boosting large-scale parallel training efficiency with c4: A communication-driven approach},
  author={Dong, Jianbo and Luo, Bin and Zhang, Jun and Zhang, Pengcheng and Feng, Fei and Zhu, Yikai and Liu, Ang and Chen, Zian and Shi, Yi and Jiao, Hairong and others},
  journal={arXiv preprint arXiv:2406.04594},
  year={2024}
}

@article{zamanzadeh2024dltsad,
  title={Deep learning for time series anomaly detection: A survey},
  author={Zamanzadeh Darban, Zahra and Webb, Geoffrey I and Pan, Shirui and Aggarwal, Charu and Salehi, Mahsa},
  journal={ACM Computing Surveys},
  volume={57},
  number={1},
  pages={1--42},
  year={2024},
  publisher={ACM New York, NY}
}

@inproceedings{hagemann2020adforcloud,
  title={A systematic review on anomaly detection for cloud computing environments},
  author={Hagemann, Tanja and Katsarou, Katerina},
  booktitle={Proceedings of the 2020 3rd Artificial Intelligence and Cloud Computing Conference},
  pages={83--96},
  year={2020}
}

@article{ding2023mstgat,
  title={MST-GAT: A multimodal spatial--temporal graph attention network for time series anomaly detection},
  author={Ding, Chaoyue and Sun, Shiliang and Zhao, Jing},
  journal={Information Fusion},
  volume={89},
  pages={527--536},
  year={2023},
  publisher={Elsevier}
}

@article{tranad,
  author= {Shreshth Tuli and
                  Giuliano Casale and
                  Nicholas R. Jennings},
  title= {TranAD: Deep Transformer Networks for Anomaly Detection in Multivariate
                  Time Series Data},
  journal      = {Proc. {VLDB} Endow.},
  volume       = {15},
  number       = {6},
  pages        = {1201--1214},
  year         = {2022},
  url          = {https://www.vldb.org/pvldb/vol15/p1201-tuli.pdf},
  doi          = {10.14778/3514061.3514067},
}

@inproceedings{audibert2020usad,
  title={Usad: Unsupervised anomaly detection on multivariate time series},
  author={Audibert, Julien and Michiardi, Pietro and Guyard, Fr{\'e}d{\'e}ric and Marti, S{\'e}bastien and Zuluaga, Maria A},
  booktitle={Proceedings of the 26th ACM SIGKDD international conference on knowledge discovery \& data mining},
  pages={3395--3404},
  year={2020}
}

@article{soldani2022adrcaformicro,
  title={Anomaly detection and failure root cause analysis in (micro) service-based cloud applications: A survey},
  author={Soldani, Jacopo and Brogi, Antonio},
  journal={ACM Computing Surveys (CSUR)},
  volume={55},
  number={3},
  pages={1--39},
  year={2022},
  publisher={ACM New York, NY}
}

@inproceedings{meng2019loganomaly,
  title={Loganomaly: Unsupervised detection of sequential and quantitative anomalies in unstructured logs.},
  author={Meng, Weibin and Liu, Ying and Zhu, Yichen and Zhang, Shenglin and Pei, Dan and Liu, Yuqing and Chen, Yihao and Zhang, Ruizhi and Tao, Shimin and Sun, Pei and others},
  booktitle={IJCAI},
  volume={19},
  number={7},
  pages={4739--4745},
  year={2019}
}

@inproceedings{xie2023gtrace,
  title={From point-wise to group-wise: A fast and accurate microservice trace anomaly detection approach},
  author={Xie, Zhe and Pei, Changhua and Li, Wanxue and Jiang, Huai and Su, Liangfei and Li, Jianhui and Xie, Gaogang and Pei, Dan},
  booktitle={Proceedings of the 31st ACM Joint European Software Engineering Conference and Symposium on the Foundations of Software Engineering},
  pages={1739--1749},
  year={2023}
}

@inproceedings{wang2023coral,
  title={Incremental causal graph learning for online root cause analysis},
  author={Wang, Dongjie and Chen, Zhengzhang and Fu, Yanjie and Liu, Yanchi and Chen, Haifeng},
  booktitle={Proceedings of the 29th ACM SIGKDD conference on knowledge discovery and data mining},
  pages={2269--2278},
  year={2023}
}

@inproceedings{guo2020gmtamicrotrace,
  title={Graph-based trace analysis for microservice architecture understanding and problem diagnosis},
  author={Guo, Xiaofeng and Peng, Xin and Wang, Hanzhang and Li, Wanxue and Jiang, Huai and Ding, Dan and Xie, Tao and Su, Liangfei},
  booktitle={Proceedings of the 28th ACM Joint Meeting on European Software Engineering Conference and Symposium on the Foundations of Software Engineering},
  pages={1387--1397},
  year={2020}
}

@inproceedings{zhang2022tracecrl,
  title={TraceCRL: contrastive representation learning for microservice trace analysis},
  author={Zhang, Chenxi and Peng, Xin and Zhou, Tong and Sha, Chaofeng and Yan, Zhenghui and Chen, Yiru and Yang, Hong},
  booktitle={Proceedings of the 30th ACM joint European software engineering conference and symposium on the foundations of software engineering},
  pages={1221--1232},
  year={2022}
}

@inproceedings{kwon2023vllm,
  title={Efficient memory management for large language model serving with pagedattention},
  author={Kwon, Woosuk and Li, Zhuohan and Zhuang, Siyuan and Sheng, Ying and Zheng, Lianmin and Yu, Cody Hao and Gonzalez, Joseph and Zhang, Hao and Stoica, Ion},
  booktitle={Proceedings of the 29th symposium on operating systems principles},
  pages={611--626},
  year={2023}
}

@inproceedings{chen2024rcacopilot,
  title={Automatic root cause analysis via large language models for cloud incidents},
  author={Chen, Yinfang and Xie, Huaibing and Ma, Minghua and Kang, Yu and Gao, Xin and Shi, Liu and Cao, Yunjie and Gao, Xuedong and Fan, Hao and Wen, Ming and others},
  booktitle={Proceedings of the Nineteenth European Conference on Computer Systems},
  pages={674--688},
  year={2024}
}

@article{wei2022cotreasoning,
  title={Chain-of-thought prompting elicits reasoning in large language models},
  author={Wei, Jason and Wang, Xuezhi and Schuurmans, Dale and Bosma, Maarten and Xia, Fei and Chi, Ed and Le, Quoc V and Zhou, Denny and others},
  journal={Advances in neural information processing systems},
  volume={35},
  pages={24824--24837},
  year={2022}
}

@inproceedings{guan2025dabl,
  title={Dabl: Detecting semantic anomalies in business processes using large language models},
  author={Guan, Wei and Cao, Jian and Gao, Jianqi and Zhao, Haiyan and Qian, Shiyou},
  booktitle={Proceedings of the AAAI Conference on Artificial Intelligence},
  volume={39},
  number={11},
  pages={11735--11744},
  year={2025}
}

@inproceedings{roy2024exploringagentforrca,
  title={Exploring llm-based agents for root cause analysis},
  author={Roy, Devjeet and Zhang, Xuchao and Bhave, Rashi and Bansal, Chetan and Las-Casas, Pedro and Fonseca, Rodrigo and Rajmohan, Saravan},
  booktitle={Companion proceedings of the 32nd ACM international conference on the foundations of software engineering},
  pages={208--219},
  year={2024}
}

@inproceedings{ahmed2023domainftllm0,
  title={Recommending root-cause and mitigation steps for cloud incidents using large language models},
  author={Ahmed, Toufique and Ghosh, Supriyo and Bansal, Chetan and Zimmermann, Thomas and Zhang, Xuchao and Rajmohan, Saravan},
  booktitle={2023 IEEE/ACM 45th International Conference on Software Engineering (ICSE)},
  pages={1737--1749},
  year={2023},
  organization={IEEE}
}

@article{prabhakar2025omniscience,
  title={OmniScience: A Domain-Specialized LLM for Scientific Reasoning and Discovery},
  author={Prabhakar, Vignesh and Islam, Md Amirul and Atanas, Adam and Wang, Yao-Ting and Han, Joah and Jhunjhunwala, Aastha and Apte, Rucha and Clark, Robert and Xu, Kang and Wang, Zihan and others},
  journal={arXiv preprint arXiv:2503.17604},
  year={2025}
}

@article{nickolls2008cuda,
  title={Scalable parallel programming with cuda: Is cuda the parallel programming model that application developers have been waiting for?},
  author={Nickolls, John and Buck, Ian and Garland, Michael and Skadron, Kevin},
  journal={Queue},
  volume={6},
  number={2},
  pages={40--53},
  year={2008},
  publisher={ACM New York, NY, USA}
}

@article{li2019nvlink,
  title={Evaluating modern gpu interconnect: Pcie, nvlink, nv-sli, nvswitch and gpudirect},
  author={Li, Ang and Song, Shuaiwen Leon and Chen, Jieyang and Li, Jiajia and Liu, Xu and Tallent, Nathan R and Barker, Kevin J},
  journal={IEEE Transactions on Parallel and Distributed Systems},
  volume={31},
  number={1},
  pages={94--110},
  year={2019},
  publisher={IEEE}
}

@article{gururangan2020dapt,
  title={Don't stop pretraining: Adapt language models to domains and tasks},
  author={Gururangan, Suchin and Marasovi{\'c}, Ana and Swayamdipta, Swabha and Lo, Kyle and Beltagy, Iz and Downey, Doug and Smith, Noah A},
  journal={arXiv preprint arXiv:2004.10964},
  year={2020}
}

@article{ouyang2022sft,
  title={Training language models to follow instructions with human feedback},
  author={Ouyang, Long and Wu, Jeffrey and Jiang, Xu and Almeida, Diogo and Wainwright, Carroll and Mishkin, Pamela and Zhang, Chong and Agarwal, Sandhini and Slama, Katarina and Ray, Alex and others},
  journal={Advances in neural information processing systems},
  volume={35},
  pages={27730--27744},
  year={2022}
}

@inproceedings{rombach2022stablediffusion,
  title={High-resolution image synthesis with latent diffusion models},
  author={Rombach, Robin and Blattmann, Andreas and Lorenz, Dominik and Esser, Patrick and Ommer, Bj{\"o}rn},
  booktitle={Proceedings of the IEEE/CVF conference on computer vision and pattern recognition},
  pages={10684--10695},
  year={2022}
}

@inproceedings{devlin2019bert,
  title={Bert: Pre-training of deep bidirectional transformers for language understanding},
  author={Devlin, Jacob and Chang, Ming-Wei and Lee, Kenton and Toutanova, Kristina},
  booktitle={Proceedings of the 2019 conference of the North American chapter of the association for computational linguistics: human language technologies, volume 1 (long and short papers)},
  pages={4171--4186},
  year={2019}
}

@inproceedings{zhang2022deeptralog,
  title={Deeptralog: Trace-log combined microservice anomaly detection through graph-based deep learning},
  author={Zhang, Chenxi and Peng, Xin and Sha, Chaofeng and Zhang, Ke and Fu, Zhenqing and Wu, Xiya and Lin, Qingwei and Zhang, Dongmei},
  booktitle={Proceedings of the 44th international conference on software engineering},
  pages={623--634},
  year={2022}
}

@inproceedings{zhang2024tracecontrast,
  title={Trace-based multi-dimensional root cause localization of performance issues in microservice systems},
  author={Zhang, Chenxi and Dong, Zhen and Peng, Xin and Zhang, Bicheng and Chen, Miao},
  booktitle={Proceedings of the IEEE/ACM 46th International Conference on Software Engineering},
  pages={1--12},
  year={2024}
}

@article{zheng2023llmjudging,
  title={Judging llm-as-a-judge with mt-bench and chatbot arena},
  author={Zheng, Lianmin and Chiang, Wei-Lin and Sheng, Ying and Zhuang, Siyuan and Wu, Zhanghao and Zhuang, Yonghao and Lin, Zi and Li, Zhuohan and Li, Dacheng and Xing, Eric and others},
  journal={Advances in neural information processing systems},
  volume={36},
  pages={46595--46623},
  year={2023}
}

@article{wang2019vfr,
  title={Rat-sql: Relation-aware schema encoding and linking for text-to-sql parsers},
  author={Wang, Bailin and Shin, Richard and Liu, Xiaodong and Polozov, Oleksandr and Richardson, Matthew},
  journal={arXiv preprint arXiv:1911.04942},
  year={2019}
}

@inproceedings{entailmentscore,
  author       = {Wojciech Kryscinski and
                  Bryan McCann and
                  Caiming Xiong and
                  Richard Socher},
  editor       = {Bonnie Webber and
                  Trevor Cohn and
                  Yulan He and
                  Yang Liu},
  title        = {Evaluating the Factual Consistency of Abstractive Text Summarization},
  booktitle    = {Proceedings of the 2020 Conference on Empirical Methods in Natural
                  Language Processing, {EMNLP} 2020, Online, November 16-20, 2020},
  pages        = {9332--9346},
  publisher    = {Association for Computational Linguistics},
  year         = {2020},
  url          = {https://doi.org/10.18653/v1/2020.emnlp-main.750},
  doi          = {10.18653/V1/2020.EMNLP-MAIN.750},
}

@article{wang2022selfconsistency,
  title={Self-consistency improves chain of thought reasoning in language models},
  author={Wang, Xuezhi and Wei, Jason and Schuurmans, Dale and Le, Quoc and Chi, Ed and Narang, Sharan and Chowdhery, Aakanksha and Zhou, Denny},
  journal={arXiv preprint arXiv:2203.11171},
  year={2022}
}

@inproceedings{lin2004rougel,
  title={Rouge: A package for automatic evaluation of summaries},
  author={Lin, Chin-Yew},
  booktitle={Text summarization branches out},
  pages={74--81},
  year={2004}
}

@inproceedings{chen2024datajuicer,
  title={Data-juicer: A one-stop data processing system for large language models},
  author={Chen, Daoyuan and Huang, Yilun and Ma, Zhijian and Chen, Hesen and Pan, Xuchen and Ge, Ce and Gao, Dawei and Xie, Yuexiang and Liu, Zhaoyang and Gao, Jinyang and others},
  booktitle={Companion of the 2024 International Conference on Management of Data},
  pages={120--134},
  year={2024}
}

@article{soldani2022cloudtsadsurvey1,
  title={Anomaly detection and failure root cause analysis in (micro) service-based cloud applications: A survey},
  author={Soldani, Jacopo and Brogi, Antonio},
  journal={ACM Computing Surveys (CSUR)},
  volume={55},
  number={3},
  pages={1--39},
  year={2022},
  publisher={ACM New York, NY}
}

@article{tan2024arellmactually,
  title={Are language models actually useful for time series forecasting?},
  author={Tan, Mingtian and Merrill, Mike and Gupta, Vinayak and Althoff, Tim and Hartvigsen, Tom},
  journal={Advances in Neural Information Processing Systems},
  volume={37},
  pages={60162--60191},
  year={2024}
}

@article{merrill2024llmstruggletotsad,
  title={Language models still struggle to zero-shot reason about time series},
  author={Merrill, Mike A and Tan, Mingtian and Gupta, Vinayak and Hartvigsen, Tom and Althoff, Tim},
  journal={arXiv preprint arXiv:2404.11757},
  year={2024}
}

@inproceedings{wang2025chattime,
  title={Chattime: A unified multimodal time series foundation model bridging numerical and textual data},
  author={Wang, Chengsen and Qi, Qi and Wang, Jingyu and Sun, Haifeng and Zhuang, Zirui and Wu, Jinming and Zhang, Lei and Liao, Jianxin},
  booktitle={Proceedings of the AAAI Conference on Artificial Intelligence},
  volume={39},
  number={12},
  pages={12694--12702},
  year={2025}
}

@inproceedings{timer,
  author       = {Yong Liu and
                  Haoran Zhang and
                  Chenyu Li and
                  Xiangdong Huang and
                  Jianmin Wang and
                  Mingsheng Long},
  title        = {Timer: Generative Pre-trained Transformers Are Large Time Series Models},
  booktitle    = {Forty-first International Conference on Machine Learning, {ICML} 2024,
                  Vienna, Austria, July 21-27, 2024},
  publisher    = {OpenReview.net},
  year         = {2024},
  url          = {https://openreview.net/forum?id=bYRYb7DMNo},
}

\end{document}